\documentclass[preprint,12pt,authoryear]{elsarticle}

\usepackage{amssymb}
\usepackage{amsmath}
\usepackage{array}      
\usepackage{caption}    
\usepackage{float}      
\usepackage{enumitem}   
\usepackage{rotating}   
\usepackage{pdflscape}  
\usepackage[utf8]{inputenc}
\usepackage[T1]{fontenc}
\usepackage{tikz}
\usepackage{algorithm}
\usepackage{algpseudocode}
\usepackage{amsmath}
\usepackage{multicol}

\usepackage{multirow}
\usepackage{float} 

\usepackage{rotating}   

\journal{Nuclear Physics B}

\begin{document}

\begin{frontmatter}



\title{Constrained Multi-Objective Genetic Algorithm Variants for Design and Optimization of Tri-Band Microstrip Patch Antenna loaded CSRR for IoT Applications: A Comparative Case Study} 


\author{Mohamed Hamza Boulaich, Said Ohamouddou, Mohammed Ali Ennasar, Abdellatif El Afia} 

\affiliation{organization={Smart System Laboratory},
            addressline={ENSIAS - Mohammed V University}, 
            city={Rabat 10120}, 
            country={Morocco}}

\begin{abstract}
This paper presents an automated antenna design and optimization framework employing multi-objective genetic algorithms (MOGAs) to investigate various evolutionary optimization approaches, with a primary emphasis on multi-band frequency optimization. Five MOGA variants were implemented and compared: the Pareto genetic algorithm (PGA), non-dominated sorting genetic algorithm with niching (NSGA-I), non-dominated sorting genetic algorithm with elitism (NSGA-II), non-dominated sorting genetic algorithm using reference points (NSGA-III), and strength Pareto evolutionary algorithm (SPEA). These algorithms are employed to design and optimize microstrip patch antennas loaded with complementary split-ring resonators (CSRRs). A weighted-sum scalarization approach was adopted within a single-objective genetic algorithm framework enhanced with domain-specific constraint handling mechanisms. The optimization addresses the conflicting objectives of minimizing the return loss ($S_{11} < -10$~dB) and achieving multi-band resonance at 2.4~GHz, 3.6~GHz, and 5.2~GHz. The proposed method delivers a superior overall performance by aggregating these objectives into a unified fitness function encompassing $S_{11}$(2.4~GHz), $S_{11}$(3.6~GHz), and $S_{11}$(5.2~GHz). This approach effectively balances all three frequency bands simultaneously, rather than exploring trade-off solutions typical of traditional multi-objective approaches. The antenna was printed on a Rogers RT5880 substrate with a dielectric constant of 2.2 , loss tangent of 0.0009 , and thickness of 1.57~mm . Scalarization approach achieved return loss values of $-21.56$~dB, $-16.60$~dB, and $-27.69$~dB, with corresponding gains of 1.96~dBi, 2.6~dB, and 3.99~dBi at 2.4~GHz, 3.6~GHz, and 5.2~GHz, respectively.
\end{abstract}


\end{frontmatter}




\section{Introduction}
Wireless communication has evolved rapidly-from 5G to 6G, and toward the Internet of Things (IoT)-based networks-driving antenna design and wireless devices into unprecedented territory \cite{ref1,ref2}. The microstrip patch antenna was first printed in 1955; however, it began receiving considerable attention in the 1970s \cite{ref3}. This antenna consists of a dielectric substrate sandwiched between a thin metallic patch and ground plane. Various geometries serve as radiating patches for microstrip antennas, including square, rectangular, and circular. In modern communication systems, rectangular patches play a crucial role due to their various advantages, including low fabrication cost, compact size, single- and multi-band frequency operation, and straightforward impedance matching. However, these structures exhibit certain limitations, such as narrow bandwidth, limited power handling capacity, and low gain \cite{ref4}. CSRRs are slot-type resonant structures that incorporate double-negative metamaterial properties, which can be engineered to exhibit unusual electromagnetic characteristics such as negative permittivity ($\varepsilon$), permeability ($\mu$), and refractive index that do not occur in natural materials. For instance, the negative refractive index causes the phase and group velocity to propagate in opposite directions, resulting energy flow and electromagnetic radiation are emitted backward instead of forward \cite{ref5}.

The design and deployment of communication systems frequently necessitate solving complex, constrained optimization problems to achieve desired performance characteristics, including high efficiency, elevated data rates, low latency, and reliable connectivity across diverse applications such as smart cities, autonomous vehicles, and medical devices. Metaheuristic algorithms aim to address these challenges and difficulties. Recently, researchers have increasingly focused on optimization algorithms, particularly evolutionary computation techniques, which efficiently balance conflicting design objectives such as minimizing antenna size while maintaining low return loss and maximizing gain. Traditional optimization methods adopt a point-by-point approach, wherein a single solution is progressively refined at each iteration, consequently yielding a single optimized solution. In contrast, evolutionary algorithms are powerful optimization tools inspired by natural selection that can solve a wide range of complex problems, particularly non-linear, multi-modal, multi-objective, and constrained optimization problems \cite{ref6}. 

 CSRRs consist of double concentric rings and squares with gaps etched into the ground plane of a microstrip patch antenna and function as quasi-static resonators. The current circulating around and within the rings generates a magnetic field that effectively behaves as an inductance, whereas the gaps and inter-ring spacing act as a capacitance. The primary challenge lies in determining the optimal position of the rings with an appropriate gap size because variations in the ring parameters and gap dimensions directly influence the resonance frequency. To address this challenge, multi-objective optimization algorithms can simultaneously minimize the antenna size, determine the optimal CSRR position, and maintain $S_{11}$ parameters below $-10$~dB at the desired frequencies. Although numerous studies have focused on single-band frequencies, multi-band frequency optimization remains relatively underexplored.

This study presents an enhanced methodology for improving the performance of microstrip patch antennas loaded with CSRRs for triple-band applications. An automated antenna design methodology was developed by integrating MATLAB with CST Microwave Studio through scripting interfaces, thereby enabling parametric optimization and simulation. A variant of the Multi-Objective Genetic Algorithm (MOGA) with constraints was employed to determine the optimal position of double rings etched in the ground plane, minimize antenna size, and maintain return loss ($S_{11}$) below the desired threshold at target frequencies of 2.4~GHz, 3.6~GHz, and 5.2~GHz.

The remainder of this paper is organized as follows. Section~2 reviews related work. Section~3 provides background information on the microstrip patch antennas with CSRRs and MOGA variants employed in this study. Section~4 describes the proposed antenna design and the implementation algorithm. Section~5 presents the performance results for the various MOGA scenarios. Section~6 concludes the paper. Finally, Section 7  discusses the future research directions. 

\section{Related Work}
Many researchers have developed various implementations of multiobjective evolutionary algorithms (MOEAs), including the pareto-based genetic algorithm (PGA) by \cite{ref7}, non-dominated sorting genetic algorithm (NSGA) by \cite{ref8}, and niched Pareto genetic algorithm (NPGA) by \cite{ref9}. These algorithms have been applied to diverse problems, demonstrating that dominance-based MOEAs can reliably identify and maintain multiple trade-offs. Genetic algorithms (GAs) have demonstrated successful application in addressing antenna design challenges \cite{ref10}. Two notable examples of antenna design employing GAs include X-band antennas for the National Aeronautics and Space Administration (NASA) Space Technology 5 spacecraft \cite{ref11} and S-band antennas for the NASA Lunar Atmosphere and Dust Environment Explorer \cite{ref12}.

~\cite{ref13} proposed a miniaturization technique for microstrip patch antennas by integrating a high-permittivity thin film as the loading element. The study employs genetic algorithms to optimize the patch dimensions, achieving a reduction in the resonance frequency from 5.8~GHz to 4.0~GHz. This optimization results in an approximately 60\% reduction in the antenna area compared with conventional designs. The optimized antenna exhibits enhanced performance characteristics, including improved return loss, bandwidth, and voltage standing wave ratio (VSWR).

~\cite{ref14} presented a multi-objective genetic algorithm (MOGA)-based optimization approach for miniaturizing hybrid compactbranch line couplers (BLCs) designed for 5G and IoT applications operating at 3.5~GHz. The MOGA framework was implemented to simultaneously balance multiple competing design objectives, specifically maintaining electrical performance while achieving significant physical size reduction. The algorithm process is automated through an interface between MATLAB, which executes the MOGA, and CST Studio Suite for electromagnetic simulation via Visual Basic for Applications (VBA) scripting. The experimental validation conducted on an FR-4 substrate demonstrated good agreement between the simulated and measured results, confirming the effectiveness of the proposed optimization methodology. 

~\cite{ref15} presented a three-dimensional (3D) electromagnetic analysis and optimization of split-ring resonators (SRRs) based on metamaterial structures. This study employed electromagnetic simulation across a frequency range of 1--20~GHz. A multi-objective genetic algorithm was used to optimize the design parameters, extending the negative refractive index region from 11~GHz to 15.5~GHz. The optimized SRR structure is suitable for wireless applications, fifth-generation (5G) antenna systems, and sensing technologies.

~\cite{ref16} presented a multi-objective optimization approach for microwave metamaterial absorbers that reduced the number of required evaluations. Rather than employing evolutionary algorithms, which require numerous objective function evaluations and are computationally expensive, the authors utilized kriging surrogate models to approximate the objective functions. The proposed algorithm was validated against benchmark problems and two well-established multi-objective optimization algorithms. Subsequently, it was successfully applied to the design of two broadband metamaterial absorbers, demonstrating both the feasibility and improved computational efficiency of the approach.

These studies contribute significantly to the design and optimization of CSRR, SRR, metamaterials, microstrip patch antennas, and couplers using multi-objective optimization. They commonly emphasize experimental validation. However, they typically employ only standard MOGAs without investigating algorithm variants, such as NSGA-II, NSGA-III, and SPEA. This research gap highlights the need to explore and compare these algorithmic variants for the design and optimization of tri-band microstrip patch antennas with CSRRs for IoT applications.

The main goal of this study is to investigate various MOGAs, including the Pareto genetic algorithm (PGA), strength Pareto evolutionary algorithm (SPEA), non-dominated sorting genetic algorithm with niching (NSGA-I), non-dominated sorting genetic algorithm with elitism (NSGA-II), and non-dominated sorting genetic algorithm using reference points (NSGA-III), to optimize antenna geometry. The optimization targets the reflection coefficient at multiple operating frequencies as well as the precise positioning and sizing of the CSRR rings etched in the ground plane. However, traditional MOGAs rely on a Pareto front with non-dominated solutions. Despite the diverse preserving mechanisms, elitism, niching, and reference points incorporated in these MOGAs, the optimal solutions consistently exhibited insufficient performance in targeted aspects, resulting in poor optimization outcomes and non-physical CSRR dimensions and positions. To address these limitations, our approach converts the multi-objective problem into a single-objective formulation by summing all three S-parameter magnitudes, thereby directly aligning with the antenna design goals to minimize the return loss at all frequencies equally.

\section{Background}
This section provides a background on MOGA and its variants. Additionally, it introduces microstrip patch antennas incorporating the CSRR in the ground plane, both of which were employed in this study. Many real-world engineering problems require simultaneous optimization of multiple conflicting objectives. Unlike single-objective optimization, which seeks a single optimal solution (i.e., a global minimum or maximum), multi-objective optimization addresses problems in which no single solution satisfies all objectives, owing to inherent trade-offs. Consequently, the goal is to identify the Pareto front set of non-dominated solutions that represents optimal trade-offs among competing objectives. MOGAs are therefore essential for antenna design, where engineers must balance conflicting performance metrics, such as the reflection coefficient, gain, bandwidth, and physical dimensions~\cite{ref8}.

\subsection{Microstrip Patch Antenna with CSRR}
Microstrip patch antennas are widely employed in modern wireless communication systems because of their low-profile geometry, lightweight construction, cost-effectiveness, and ease of integration with electronic circuits. A microstrip patch antenna comprises a thin metallic patch printed on the upper surface of a dielectric substrate with a metallic ground plane positioned on the lower surface. The radiating patch can assume various geometries, including rectangular, circular, and triangular. A CSRR is a metamaterial-inspired structure etched into a metallic surface, consisting of one or more concentric rings with narrow gaps that form resonant slots. The CSRR exhibits negative permittivity ($\varepsilon$-negative) resonant behavior~\cite{ref17}.

\subsection{Pareto-based Genetic Algorithm}
The pareto-based evolutionary algorithm was first introduced by~\cite{ref45}, who proposed assigning equal reproductive probabilities to all non-dominated individuals within the population. His approach involved assigning rank 1 to non-dominated individuals before removing them from consideration, subsequently identifying the next set of non-dominated individuals, and assigning them rank 2, with this process continuing iteratively. Fonseca and Fleming later extended this methodology by proposing a ranking system in which an individual's rank is determined by the number of population members that dominate. Under their scheme, all non-dominated individuals receive an identical rank, whereas dominated individuals are penalized based on the population density within the corresponding region of the trade-off surface~\cite{ref18}. Instead of assigning fitness directly to the objective values, a Pareto ranking mechanism based on dominance count was proposed. Fonseca and Fleming considered an individual $x_i$ at generation $t$ which was dominated by $n_d(x_i)$ individuals in the current population. Its rank can be given by~\cite{ref7}.

\begin{equation}
r(x_i) = 1 + n_d(x_i)
\end{equation}

where:
\begin{equation}
n_d(x_i) = \left|\{x_j \in P \mid x_j \prec x_i\}\right|
\end{equation}

\noindent In the field of electronics and communications, numerous studies have applied Pareto-based optimization techniques, such as Pareto genetic algorithms (PGAs) and other multi-objective evolutionary algorithms, to optimize multiple conflicting objectives. For instance, ~\cite{ref19} applied a Pareto genetic algorithm to optimize the design of a broadband planar dipole antenna for a long wavelength array (LWA) and a low-frequency radio telescope. The algorithm simultaneously optimized the antenna radiation patterns and minimized the received galactic background noise while respecting the design constraints to ensure realizability and cost control.

In our case, a Pareto-based approach following the Fonseca and Fleming framework with elitism and fitness sharing was applied to classify the solutions according to dominance. Fitness sharing is incorporated to preserve the population diversity and prevent premature convergence. Elitism and tournament selection were employed to retain high-quality solutions, followed by crossover and mutation operators to explore the search space. The final outcome is a Pareto front consisting of non-dominated antenna designs that represent optimal trade-offs among the considered frequency bands. Algorithm 1 presents the Pareto-based genetic algorithm used to optimize the antenna design.

\begin{algorithm*}
\caption{Pareto-based Genetic Algorithm for Antenna Optimization}
\footnotesize
\begin{algorithmic}[1]
\State \textbf{Input:} Population size $N$, generations $G$, design constraints
\State \textbf{Output:} Pareto front of non-dominated solutions (optimal antenna geometries)
\State Initialize antenna parameters: $R_1, R_2, R_3, R_4, V_r, U_r, W_g, L_g, W, L$
\State Initialize population $P$ with antenna parameters
\For{$g=1$ to $G$}
    \ForAll{individuals in $P$}
        \State Evaluate EM model and compute $S_{11}$ at 2.4, 3.6, and 5.2 GHz
    \EndFor
    \State Apply Pareto ranking (Fonseca--Fleming)
    \State Apply fitness sharing
    \State Extract rank-1 solution
    \State selection (elitism + tournament)
    \State Apply crossover and mutation
    \State Update population $P$
\EndFor
\State \Return Pareto front.
\end{algorithmic}
\end{algorithm*}

\subsection{Non-dominated Sorting Genetic Algorithm with Niched (NSGA-I).}
The non-dominated sorting genetic algorithm (NSGA-I) is a multi-objective optimization algorithm that combines the non-dominated sorting principle with niching techniques. This algorithm applies niching mechanisms through fitness sharing to maintain the population diversity across the Pareto front. One of the most widely implemented selection techniques for genetic algorithms is tournament selection, in which a subset of individuals is randomly chosen from the current population, and the fittest individual from this subset is selected for the next generation. Tournament selection tends to drive the population toward a single solution over successive generations, thereby causing premature convergence. To maintain multiple Pareto-optimal solutions, NSGA employs a selection mechanism based on non-dominated sorting, and fitness sharing. First, individuals were classified into different non-dominated fronts through non-dominated sorting. Second, a sharing mechanism was applied within each front to maintain diversity by reducing the fitness of individuals in crowded regions. Proportionate selection is then performed based on these shared fitness values. This approach prevents premature convergence and preserves the diversity along the Pareto front~\cite{ref8}. The niche count $n_i$ quantifies the degree of crowding within the neighborhood (niche) surrounding individual $i$. This metric is computed by evaluating all individuals present in the current population.

\begin{equation}
n_i = \sum_{j=1}^{n} sh(d_{ji})
\end{equation}

where:

$d_{ji}$: is the distance between individuals $i$ and $j$

$sh(d)$: is the sharing function, which maintains diversity:

\begin{equation}
sh(d_{ji}) = 
\begin{cases}
\displaystyle 1 - \left(\frac{d_{ji}}{\sigma_{share}}\right)^{\alpha}, & d_{ji} < \sigma_{share} \\[0.3cm]
0, & d_{ji} \geq \sigma_{share}
\end{cases}
\end{equation}

where:

$\sigma_{share}$: Niche radius

$\alpha$: Sharing parameter

The rapid growth of wireless communication technology has led to increased reliance on antennas in modern vehicles. Designing these antennas for optimal performance is challenging because it often requires balancing multiple conflicting objectives, such as radiation pattern characteristics, bandwidth, and physical constraints.~\cite{ref20} presented a computational design approach for conformal vehicle antennas using NSGA. This method optimizes antenna geometries to achieve an improved multi-objective performance. The approach was validated through electromagnetic simulations and experimental measurements, thereby demonstrating its effectiveness in vehicle antenna design applications. 

In this study, NSGA-I evaluates each candidate design using three objectives and assigns ranks through non-dominated sorting. To preserve the population diversity and prevent convergence to a single region of the objective space, a niching mechanism based on fitness sharing is employed, which reduces the fitness of closely clustered solutions. This mechanism ensures a well-distributed set of solutions along the Pareto front, thereby maintaining the exploration capability of the algorithm throughout the optimization process. Algorithm 2 presents the NSGA-I used to optimize the antenna design.

\begin{algorithm}
\caption{Non-dominated Sorting Genetic Algorithm with Niched (NSGA-I) for Antenna Optimization}
\begin{algorithmic}[1]
\State \textbf{Input:} Population size $N$, Generations $G$, Parameter constraints
\State \textbf{Output:} Pareto Front (Rank 1 solutions), (optimal antenna geometries)
\State Initialize antenna parameters: $R_1, R_2, R_3, R_4, V_r, U_r, W_g, L_g, W, L$
\State Initialize population $P$ with antenna parameters

\For{$g = 1$ to $G$}
    \State Evaluate objectives ($S_{11}$ at 2.4, 3.6, 5.2 GHz) for all individuals
    \State Perform \textbf{Non-Dominated Sorting} and assign ranks
    \State Apply \textbf{Fitness Sharing} to maintain diversity
    \State Extract Pareto Front (Rank 1)
    \State Apply \textbf{Tournament Selection}, \textbf{Crossover}, and \textbf{Mutation} to create next generation
\EndFor

\State \textbf{return} Pareto Front
\end{algorithmic}
\end{algorithm}

\subsection{Non-dominated Sorting Genetic Algorithm with Elitism (NSGA-II)}
The Elitist Non-dominated Sorting Genetic Algorithm (NSGA-II) is an improved multi-objective optimization algorithm that extends NSGA by incorporating elitism to preserve the best solutions across generations. This algorithm employs a fast non-dominated sorting procedure to classify individuals into different fronts based on dominance relationships. Two approaches exist for sorting a population into different non-domination levels: naive and efficient approach. The naive approach compares each solution against all others in the population with a computational complexity of $O(MN^2)$, where $N$ is the population size, and $M$ is the number of objectives. To identify all members of the first non-dominated front, this comparison must be performed for each individual, resulting in $O(MN^2)$ complexity in identifying the first front. To identify individuals in subsequent fronts, the solutions in the first front were temporarily removed, and the sorting procedure was repeated for the remaining population. Identifying the second front also requires $O(MN^2)$ comparisons, particularly when $O(N)$ solutions belong to the second non-dominated level. This approach is sufficient for identifying all non-domination levels. In the worst case, when there are $N$ fronts with only one solution in each front, the overall complexity is $O(MN^3)$~\cite{ref21}. Multi-objective evolutionary algorithms have been applied in the field of satellite communications to optimize broadband reflector antenna designs.~\cite{ref22} proposed a Non-dominated Sorting Genetic Algorithm based on Reinforcement Learning (NSGA-RL), which is an improved version of the NSGA-II that integrates a parameter-free self-tuning approach using reinforcement learning. NSGA-RL was applied to a satellite coverage optimization problem and compared with the classical NSGA-II and other multi-objective optimization methods. 

In the present study, NSGA-II evaluates antenna geometry by incorporating CSRR structures and ground/patch dimensions through full-wave electromagnetic simulations. The algorithm applies fast non-dominated sorting to classify solutions according to Pareto dominance principles. Elitism is ensured by combining the parent and offspring populations and selecting the best solutions based on their dominance rank. This mechanism enables the algorithm to converge toward a well-distributed Pareto front that represents the optimal trade-offs among the targeted operating frequencies. Algorithm 3 presents the NSGA-II used to optimize the antenna design.

\begin{algorithm}
\caption{Non-dominated Sorting Genetic Algorithm with Elitism (NSGA-II) for Antenna Optimization}
\begin{algorithmic}[1]
\State \textbf{Input:} Population size $N$, Generations $G$, Parameter constraints
\State \textbf{Output:} Pareto Front (optimal antenna geometries)
\State Initialize antenna parameters: $R_1, R_2, R_3, R_4, V_r, U_r, W_g, L_g, W, L$
\State Initialize population $P$ with antenna parameters
\For{$g = 1$ to $G$}
    \State Evaluate objectives ($S_{11}$ at 2.4, 3.6, 5.2 GHz)
    \State Apply \textbf{Fast Non-Dominated Sorting}
    \State Generate offspring via \textbf{Selection, Crossover, Mutation}
    \State Combine parent and offspring populations
    \State Select best $N$ individuals using \textbf {rank} 
\EndFor
\State \textbf{return} Pareto Front
\end{algorithmic}
\end{algorithm}

\subsection{Non-dominated Sorting Genetic Algorithm with Reference Points (NSGA-III)}
NSGA-III extends NSGA-II by introducing a reference-point-based selection mechanism, representing a significant modification to the original algorithm for multi-objective optimization. Although the crowding distance mechanism in NSGA-II effectively maintains diversity in bi- and tri-objective problems, it becomes ineffective when addressing problems with a large number of objectives. NSGA-III overcomes this limitation by employing a predefined set of reference points to guide the population toward a well-distributed Pareto front. This approach achieves improved diversity, enhanced uniform coverage of high-dimensional objective spaces, and better convergence behavior, making it particularly suitable for multi-objective problems with four or more objectives.

The NSGA-III ensures solution diversity through a predetermined set of reference points distributed on a normalized hyperplane. These reference points can be either user--specified or systematically generated using structured placement strategies. The reference points are positioned on an $(M-1)$-dimensional unit simplex with unit intercepts on each axis, maintaining an equal inclination to all objective axes. For an $M$-objective optimization problem, the total number of reference points ($H$) is determined based on $p$ divisions along each objective axis~\cite{ref23}:

\begin{equation}
H = \binom{M + p - 1}{p}
\end{equation}

For example, in a three-objective problem ($M = 3$), reference points are created on a triangle with vertices at $(1,0,0)$, $(0,1,0)$, and $(0,0,1)$. If four divisions ($p = 4$) are selected for each objective axis.

\begin{equation}
H = \binom{3 + 4 - 1}{4}
\end{equation}

Frequency-selective surface design presents significant optimization challenges, and NSGA-III has been widely adopted to address these complexities.~\cite{ref24} proposed a systematic inverse-design method for miniaturizing frequency-selective surfaces (FSSs) using microwave network theory and equivalent circuit models. A multi-objective optimization problem was formulated and solved using NSGA-III. The method successfully generated three highly miniaturized FSS designs: single-band transmissive FSS, single-band reflective FSS, and a multi-band reflective FSS. 

In this study, NSGA-III applies non-dominated sorting and utilizes a set of predefined reference points to guide the environmental selection process, thereby ensuring a well-distributed Pareto front in the objective space. Algorithm~4 presents the NSGA-III procedure employed to optimize the antenna design.

\begin{algorithm}
\caption{Non-dominated Sorting Genetic Algorithm with Reference Points (NSGA-III) for Antenna Optimization}
\begin{algorithmic}[1]
\State \textbf{Input:} Population size $N$, generations $G$, design constraints
\State \textbf{Output:} Pareto-optimal antenna solutions
\State Initialize antenna parameters: $R_1, R_2, R_3, R_4, V_r, U_r, W_g, L_g, W, L$
\State Generate reference points using Das--Dennis method
\State Initialize population $P$ with antenna parameters

\For{$g = 1$ to $G$}
    \State Evaluate population using EM simulation
    \State Compute objectives: $S_{11}$ at 2.4, 3.6, and 5.2 GHz
    \State Perform non-dominated sorting
    \State Generate offspring via crossover and mutation
    \State Combine parent and offspring populations
    \State Update population $P$
\EndFor

\State \textbf{return} Pareto Front (Rank-1 solutions)
\end{algorithmic}
\end{algorithm}

\subsection{Strength Pareto Evolutionary Algorithm (SPEA)}
The strength Pareto evolutionary algorithm (SPEA) is a multi-objective optimization technique designed to identify high-quality Pareto solutions. SPEA combines traditional evolutionary algorithm methods, such as mutation, crossover, and selection, which are common techniques employed in most genetic algorithms. It is specifically designed for multi-objective optimization and incorporates an external archive that stores the best non-dominated solutions~\cite{ref25} and a strength fitness assignment mechanism, where each solution receives a score based on the number of other solutions it dominates~\cite{ref26}.

In the present study, the SPEA maintained an external archive to store non-dominated solutions, ensuring the preservation of Pareto-optimal designs throughout the evolutionary process. Each individual in the combined population and archive is assigned a strength and fitness value based on

\begin{algorithm}
\caption{Strength Pareto Evolutionary Algorithm (SPEA) for Antenna Optimization}
\begin{algorithmic}[1]
\State \textbf{Input:} Population size $N$, archive size $\bar{N}$, generations $G$
\State \textbf{Output:} External archive (Pareto Front)

\State Initialize population $P$ and external archive $\bar{P}$

\For{$g = 1$ to $G$}
    \State Evaluate antenna designs using EM simulation
    \State Compute objectives: $S_{11}$ at 2.4, 3.6, and 5.2 GHz
    \State Combine population and archive
    \State Assign strength and fitness values
    \State Update external archive with non-dominated solutions
    \State Apply selection, crossover, and mutation
    \State Update population $P$
\EndFor

\State \textbf{return} External archive $\bar{P}$
\end{algorithmic}
\end{algorithm}

\noindent dominance relationships. This approach ultimately produces a set of diverse Pareto-optimal antenna configurations that balance performance across all target frequencies. Algorithm~5 presents the SPEA procedure employed to optimize the antenna design.

Table 1 presents a comparative review of five widely employed MOGA variants applied in antenna design and optimization, namely Pareto GA, NSGA-I, NSGA-II, NSGA-III, and SPEA. The comparison highlights their strengths and limitations with respect to the essential criteria for electromagnetic optimization, including diversity preservation, elitism, and suitability for multi-objective antenna scenarios.

\section{The Proposed Approach}
In brief, we developed an automated methodology for designing a microstrip patch antenna with a CSRR that features a gap etched in the ground plane. The design and optimization process were implemented using MATLAB-CST integration and constrained MOGAs. The optimization objectives included determining the optimal CSRR position and achieving size reduction while maintaining a reflection coefficient below $-10$ dB at the target frequencies of 2.4 GHz, 3.6 GHz, and 5.2 GHz. Five variants of multi-objective GAs were investigated: PGA, NSGA-I, NSGA-II, NSGA-III, and SPEA. We adopt a weighted-sum scalarization approach within a single-objective framwork to address the limitations and challenges inherent in these algorithms.

\subsection{System Architecture}
As depicted in figure 1, our architecture integrates two distinct software components, MATLAB and CST. Each component fulfills a specific role within the workflow. MATLAB serves as the optimization and data preprocessing engine for the antenna design process, whereas CST functions as a full-wave electromagnetic simulator. The Application Programming Interface API enables MATLAB to control CST automatically, facilitating seamless integration between the two platforms.

\begin{table}[H]
\makebox[\textwidth][c]{
\begin{minipage}{1.2\textwidth}  
\centering
\caption{Well-known multi-objective genetic algorithm variants}
\label{tab:multi-objective-ga}
\footnotesize
\setlength{\tabcolsep}{4pt}
\renewcommand{\arraystretch}{0.9}
\begin{tabular}{|>{\centering\arraybackslash}m{1.7cm}|>{\centering\arraybackslash}m{2cm}|>{\centering\arraybackslash}m{2cm}|>{\centering\arraybackslash}m{1.2cm}|>{\centering\arraybackslash}m{1.3cm}|>{\centering\arraybackslash}m{1.5cm}|m{2.8cm}|m{2.8cm}|}
\hline
\textbf{Ref} & \textbf{Algorithm} & \textbf{Fitness} & \textbf{Elitism} & \textbf{Niching} & \textbf{External archive} & \textbf{Advantage} & \textbf{disadvantage} \\
\hline
~\cite{ref27} & PGA & Rank based on level of Pareto dominance & NO & NO & NO & 
$\bullet$ Low computational cost \newline
$\bullet$ Clear objective \newline
$\bullet$ Multi-objective optimization & 
$\bullet$ Method details lacking \newline
$\bullet$ Complex terminology \newline
$\bullet$ No quantitative results \\
\hline
~\cite{ref28} & NSGA-I & Pareto rank + sharing & NO & Yes & NO & 
$\bullet$ Fast convergence & 
$\bullet$ Problems related to niche size parameter \\
\hline
~\cite{ref29} & NSGA-II & Pareto rank + crowding distance & Yes & NO & NO & 
$\bullet$ Fast convergence \newline
$\bullet$ Efficient multi-objective optimization \newline
$\bullet$ Good diversity in Pareto solutions & 
$\bullet$ Requires parameter tuning for NSGA-II \newline
$\bullet$ Validation sometimes limited to simulation \newline
$\bullet$ Computational cost for moderate\\
\hline
~\cite{ref30} & NSGA-III & Pareto rank + reference point & YES & NO & NO & 
$\bullet$ Reduced computational cost \newline
$\bullet$ Reliable and efficient solution \newline
$\bullet$ Handles multi-dimensional parameters well & 
$\bullet$ Risk of convergence to local minima\newline
$\bullet$ Difficult to handle very high-dimensional problems \newline
$\bullet$ Performance depends on simulator accuracy \\
\hline
~\cite{ref31} & SPEA & Strength value (dominance count) & YES & NO & YES & 
$\bullet$ Good sidelobe suppression \newline
$\bullet$ Suitable for random/irregular arrays \newline
$\bullet$ Efficient Pareto approximation & 
$\bullet$ Higher computational cost \newline
$\bullet$ Parameter tuning required \newline
$\bullet$ Slower convergence \\
\hline
\end{tabular}
\end{minipage}
}
\end{table}

\noindent The workflow comprises six sequential stages.

\textbf{First}, candidate antenna geometries with CSRRs are generated, including parameters such as ring radii ($R_1$, $R_2$, $R_3$, $R_4$), gap sizes ($g_1$, $g_2$), patch dimensions ($W_g$, $L_g$, $W_p$, $L_p$), and position variables ($V_r$, $U_r$). as well as the substrate, patch, and ground plane material properties and the associated constraints and objective functions.

\textbf{Second}, a genetic algorithm and multi-objective optimization methods are implemented, followed by the evaluation of fitness functions, including $S_{11}$ performance at multiple frequencies, bandwidth, and constraint satisfaction.

\textbf{Third}, the geometric parameters were created inside the CST through the API.

\textbf{Fourth}, the electromagnetic simulation engine in CST computes all electromagnetic characteristics of the antenna.

\textbf{Fifth}, data retrieval and post-processing were performed to extract S-parameters, frequency responses, and current distributions from the CST output files (ASCII tables), and performance indicators ($S_{11}$, gain) were computed.

\textbf{Finally}, decision-making for optimization is executed by selecting the best geometry according to the specified constraints and design objectives.

\begin{figure*}[t!]
\centering
\includegraphics[width=1\textwidth]{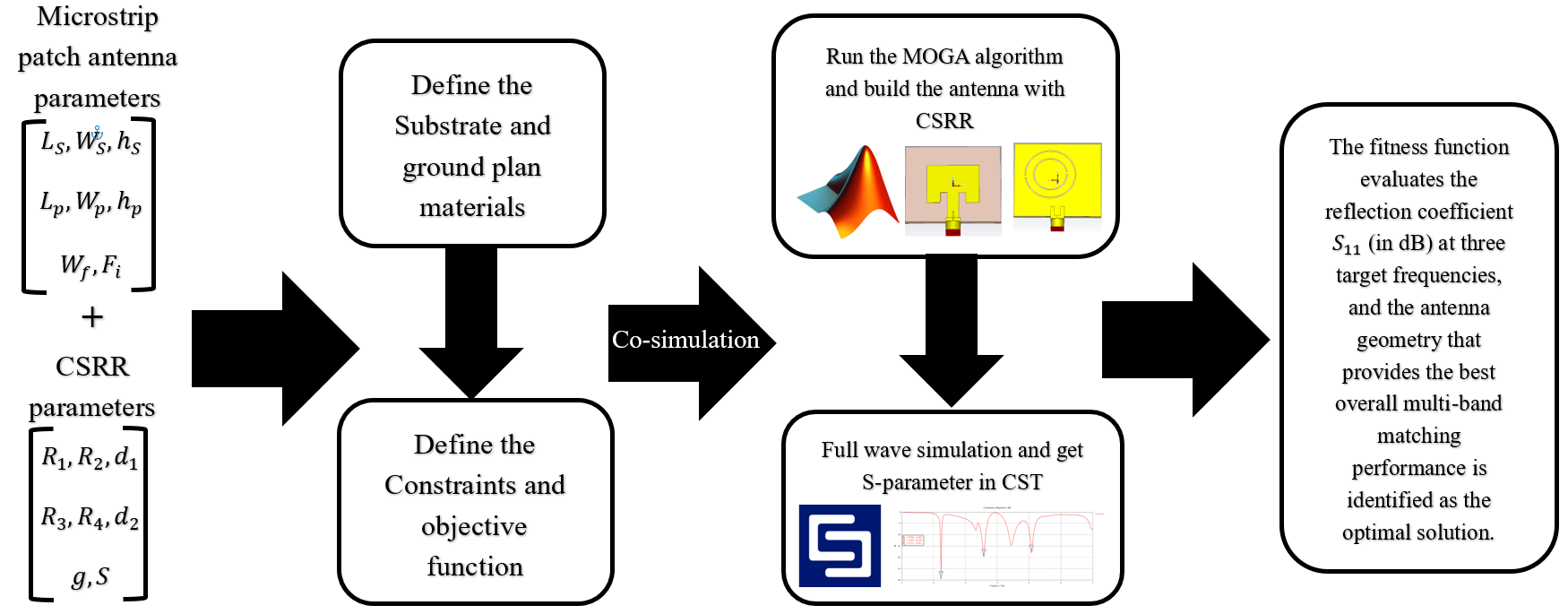}
\caption{The architecture of our proposed approach}
\label{fig1}
\end{figure*}

\subsection{Design and Implementation.}
\subsubsection{Antenna Structure and Configuration}

First, we present a rectangular microstrip patch antenna operating at 5.2 GHz , which is selected as the reference antenna for the initial design. The primary physical dimensions of the reference antenna included the width, length, and thickness of the radiation patch: $W_p = 22.8$ mm, $L_p = 18.55$ mm , and $h_p = 35$ $\mu$m, respectively. These dimensions determine the fundamental operating frequency and impedance behavior of the antenna. The antenna was printed on a dielectric substrate characterized by a relative permittivity ($\varepsilon_r = 2.2$) and loss tangent ($\tan \delta = 0.0009$) corresponding to the Rogers RT5880 material. The dielectric constant controls the miniaturization level and electromagnetic confinement within the structure, whereas the loss tangent quantifies the material losses and influences the antenna efficiency. 

A microstrip feed line with dimensions $W_f = 4.85$ mm and $L_f = 14.3$ mm was implemented to achieve a $50~\Omega$ characteristic impedance, providing proper impedance matching of the antenna. The microstrip feed line is inset toward the patch at $D = 5.02$ mm and is separated by two gaps of equal width $S = 3.1$ mm. Figure 2 (a) shows the antenna structure and Figure 2 (b) shows the antenna resonating at 5.2 GHz.

\begin{figure*}[t!]
\centering
\includegraphics[width=1\textwidth]{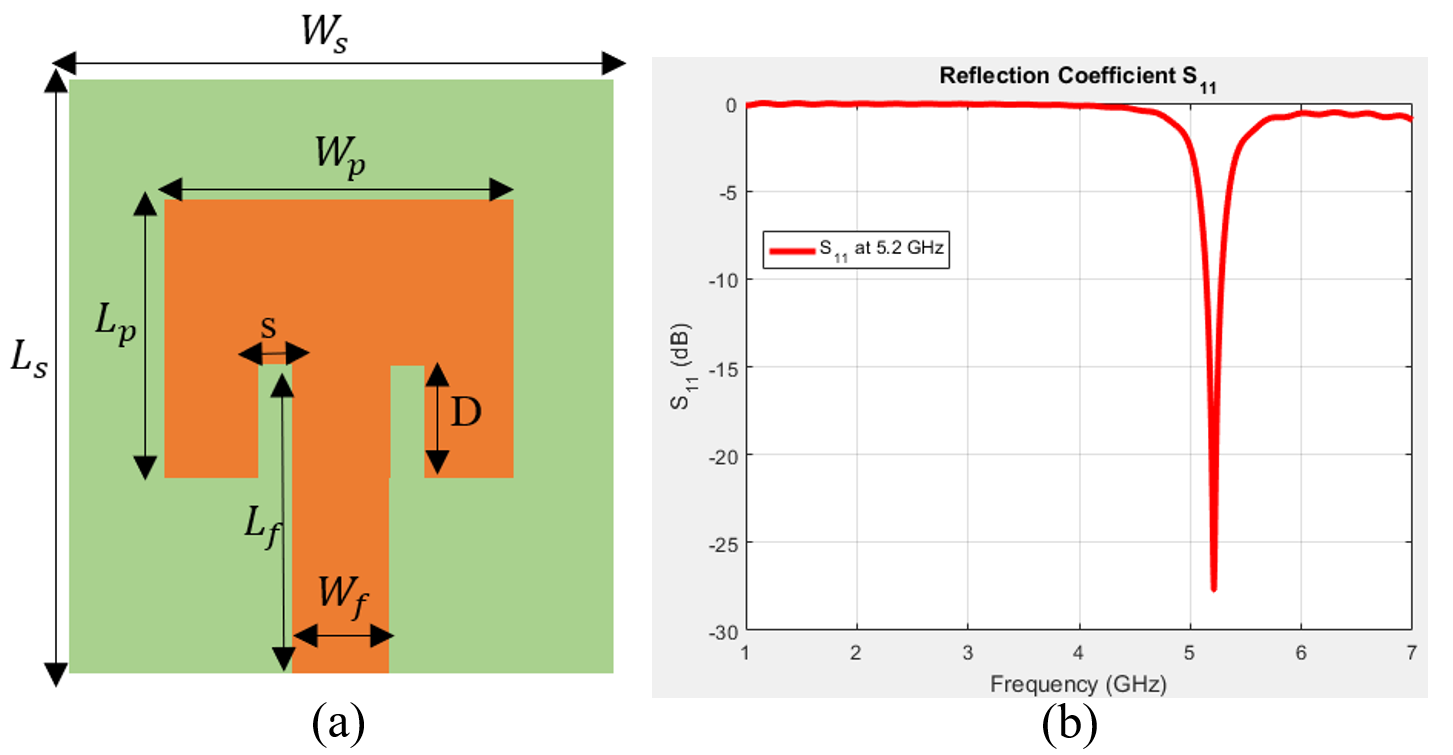}
\caption{Reference microstrip patch antenna at 5.2 GHz (a) Geometry, (b) Reflection Coefficient $S_{11}$}
\label{fig1}
\end{figure*}

\subsubsection{Development of the Microstrip Patch Antenna}
The proposed antenna is developed using a multi-stage design process. The two principal stages are illustrated in Figure 3(a) and (b). In the first stage, an initial ring with a gap is etched and implemented as a defective ground structure (DGS) to introduce a new resonant frequency at 2.4~GHz. The antenna performance was further enhanced by etching the CSRR structure into the ground plane. The CSRR position was determined not by the current distribution within the ground plane, but rather by a constrained MOGA. Furthermore, the CSRR dimensions were optimized using this algorithm to achieve resonant frequencies within the desired frequency bands while maintaining reflection coefficient $S_{11}$ below $-10$~dB. In the second stage, the objective was to introduce an additional resonant frequency of 3.6~GHz, thereby enabling tri-band resonant operation. To achieve this objective, a second CSRR with a gap is designed and incorporated. This modification enables the antenna to exhibit enhanced multiband characteristics and improved performance without increasing its physical dimensions. The parametric studies conducted throughout these stages demonstrated the influence of the number of rings on the generation of the resonance frequencies and the corresponding $S_{11}$ performance. The initial dimensions of the proposed antenna prior to the MOGA optimization, calculated using mathematical formulations, are listed in Table~2.

\begin{table}[H]
\centering
\caption{Units and values of the proposed antenna parameters.}
\label{tab:antenna-parameters}
\begin{tabular}{cccc}
\hline
Parameter & Value (mm) & parameter & Value (mm) \\
\hline
$W_S$ & 41.64 & $R_1$ & 12.38 \\
$L_S$ & 37.93 & $R_2$ & 11.88 \\
$h$ & 1.57 & $R_3$ & 8.25 \\
$W_p$ & 22.8 & $R_4$ & 7.75 \\
$L_p$ & 18.55 & $d_1$ & 0.5 \\
$L_f$ & 14.3 & $d_2$ & 0.5 \\
$W_f$ & 4.85 & $s$ & 5 \\
$D$ & 5.02 & $gap$ & 1.5 \\
$S$ & 3.1 & $\varepsilon_r$ & 2.2 \\
\hline
\end{tabular}
\end{table}

\begin{figure*}[t!]
\centering
\includegraphics[width=1\textwidth]{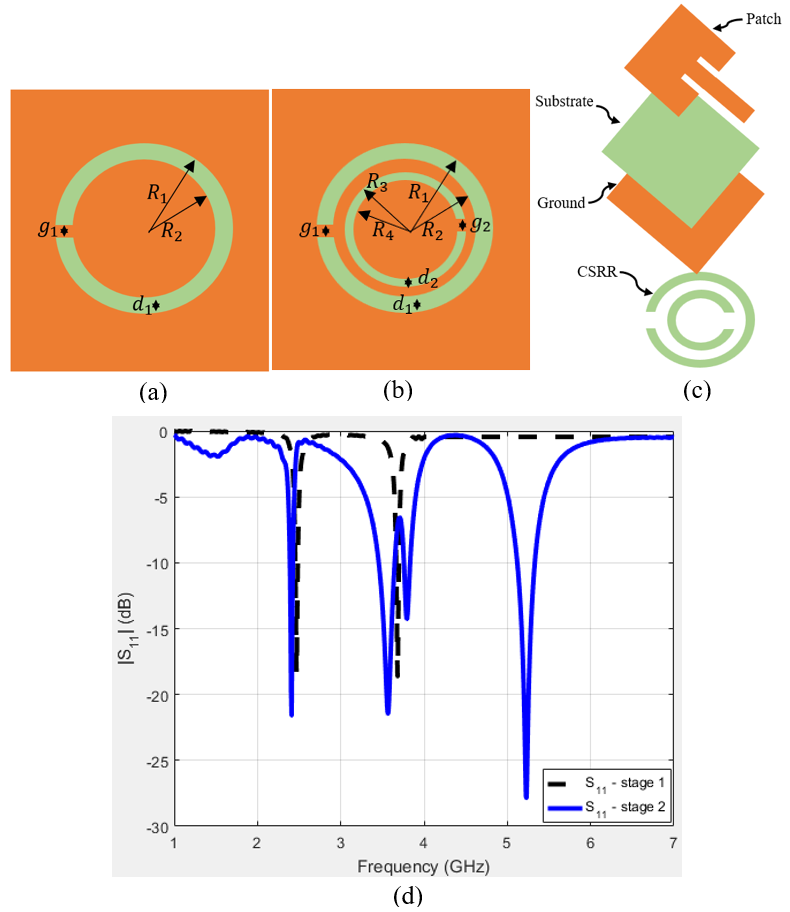}
\caption{Evolution of the microstrip patch antenna with CSRR: (a)~Stage~1, (b)~Stage~2, (c)~schematic representation of the proposed multiband antenna, and (d)~reflection coefficient $S_{11}$ for the various design stages.}
\label{fig1}
\end{figure*}

\subsection{Proposed Algorithm}
In this study, a modified genetic algorithm was developed to converts the multi-objective antenna optimization problem into a single-objective formulation using weighted-sum scalarization, enhanced with constraint-repair mechanisms and elitist preservation strategies. In addition, various MOGAs have been investigated, including PGA, NSGA-I, NSGA-II, NSGA-III, and SPEA. These algorithms were applied to optimize antenna geometry. The optimization process targeted the position and dimensions of the double rings with gaps in the CSRR structure, aiming to reduce the antenna size while maintaining the reflection coefficient $S_{11}$ below $-10$~dB at specific frequencies of 2.4~GHz, 3.6~GHz, and 5.2~GHz.

The proposed algorithm addresses the limitations of previous multiobjective algorithms in achieving optimal CSRR positioning within the ground

\noindent
\begin{minipage}[t]{0.48\textwidth}
\centering
\footnotesize

\usetikzlibrary{shapes.geometric, arrows, positioning, arrows.meta, calc}
\tikzset{
    startstop/.style={rectangle, rounded corners, minimum width=2cm, minimum height=0.7cm, text centered, draw=black, fill=green!30},
    process/.style={rectangle, minimum width=2.5cm, minimum height=0.7cm, text centered, draw=black, fill=orange!30},
    decision/.style={diamond, minimum width=2cm, minimum height=2cm, text centered, draw=black, fill=yellow!30, inner sep=1pt},
    fitness/.style={rectangle, minimum width=2.5cm, minimum height=0.7cm, text centered, draw=black, fill=blue!30},
    simulation/.style={rectangle, minimum width=2.5cm, minimum height=0.7cm, text centered, draw=black, fill=violet!30},
    arrow/.style={->,>=stealth}
}
\clearpage
\begin{tikzpicture}[node distance=1.2cm, scale=0.85, transform shape]
\node (start) [startstop] {START};
\node (init) [process, below of=start,yshift=-0.1cm, text width=5cm] {Initialize antenna parameters \\$W, L, W_f, L_g, W_g,$\\$R_1, R_2, R_3, R_4, U_r, V_r$};
\node (initpop) [process, below of=init,yshift=-0.2cm, text width=5cm] {Initialize Population\\$N = 10$};
\node (genloop) [decision, below of=initpop, yshift=-1.1cm] {Generation 1 to 10};
\node (evaluate) [process, below of=genloop, yshift=-1cm] {Evaluate Individuals $i = 1$ to 10};
\node (runsim) [simulation, below of=evaluate, yshift=-0.3cm, text width=4cm] {Run Simulation\\myFunction\\$(W, L, W_f, L_g, W_g,$\\$R_1, R_2, R_3, R_4, U_r, V_r)$};
\node (objectives) [process, below of=runsim, yshift=-0.4cm, text width=4cm] {3 Objectives\\$S_{11}$ at 2.4, 3.6, 5.2 GHz};
\node (fitness) [fitness, below of=objectives, yshift=-0.2cm, text width=4cm] {Calculate Fitness\\ $F = |S_{11,2.4}| + |S_{11,3.6}| + |S_{11,5.2}|$};
\node (alleval) [decision, below of=fitness, yshift=-0.5cm] {All Done?};
\node (findbest) [fitness, below of=alleval, yshift=-0.7cm, text width=2.5cm] {Find Best Individual\\(Max Fitness)};
\node (updatebest) [process, below of=findbest,yshift=-0.3cm, text width=2.5cm] {Update Overall\\Best Solution};
\node (checklast) [decision, below of=updatebest, yshift=-0.5cm] {Last Gen?};
\node (output) [process, below of=checklast, yshift=-0.6cm, text width=2.5cm] {Output Best\\Solution};
\node (elitism) [process, right of=checklast, xshift=2cm, text width=2.5cm] {Elitism\\Keep Best Individual};
\node (selection) [process, below of=elitism,yshift=-0.2cm,  text width=2.5cm] {Tournament Selection};
\node (genetic) [process, below of=selection, text width=2.5cm] {Crossover\\+ Mutation};
\node (newpop) [process, below of=genetic] {New Population};
\node (end) [startstop, below of=output] {END};
\draw [arrow] (start) -- (init);
\draw [arrow] (init) -- (initpop);
\draw [arrow] (initpop) -- (genloop);
\draw [arrow] (genloop) -- (evaluate);
\draw [arrow] (evaluate) -- (runsim);
\draw [arrow] (runsim) -- (objectives);
\draw [arrow] (objectives) -- (fitness);
\draw [arrow] (fitness) -- (alleval);
\draw [arrow] (alleval) -- node[anchor=east, font=\tiny] {Yes} (findbest);
\draw [arrow] (alleval) -| node[anchor=south, pos=0.25, font=\tiny] {No} ++(-2.8,0) |- (evaluate);
\draw [arrow] (findbest) -- (updatebest);
\draw [arrow] (updatebest) -- (checklast);
\draw [arrow] (checklast) -- node[anchor=east, font=\tiny] {Yes} (output);
\draw [arrow] (checklast) -- node[anchor=south, font=\tiny] {No} (elitism);
\draw [arrow] (elitism) -- (selection);
\draw [arrow] (selection) -- (genetic);
\draw [arrow] (genetic) -- (newpop);
\draw [arrow] (newpop) -- ++(0,-0.5) -| ++(-6.5,0) |- (genloop);
\draw [arrow] (output) -- (end);
\end{tikzpicture}

\vspace{0.3em}
\textbf{Flowchart of the Scalarized MOGA Optimization}
\end{minipage}
\hfill
\begin{minipage}[t]{0.7\textwidth}
\footnotesize

\begin{algorithm}[H]
\caption{Single-Objective (Scalarized Multi-Objective Genetic Algorithm) for Antenna Optimization}
{\scriptsize  
\begin{algorithmic}[1]
\State \textbf{Input:} Population size $N=10$, Generations $G=10$, Parameters constraints, antenna parameters: $R_1, R_2, R_3, R_4, V_r, U_r, W_g, L_g, W, L$
\State \textbf{Output:} Best solution (optimal antenna parameters)
\State Initialize antenna parameters: $R_1, R_2, R_3, R_4, V_r, U_r, W_g, L_g, W, L$
\State Initialize population $P$ with $N$ individuals
\State $bestSolution \gets \emptyset$, $bestFitness \gets -\infty$
\For{$g = 1$ to $G$}
    \For{$i = 1$ to $N$}
        \State Run simulation: $\text{myFunction}(R_1, R_2, R_3, R_4, V_r, U_r, W_g, L_g, W, L)$
        \State Calculate 3 objectives: $S_{11}$ at 2.4, 3.6, 5.2 GHz
        \State \textbf{Scalarization (MOGA approach):}
        \State \quad $F_i = |S_{11,1}| + |S_{11,2}| + |S_{11,3}|$ (aggregate multiple objectives)
    \EndFor
    
    \State Find best individual: $best_g \gets \arg\max_i F_i$
    
    \If{$F_{best_g} > bestFitness$}
        \State Update overall best: $bestSolution \gets P[best_g]$
        \State $bestFitness \gets F_{best_g}$
    \EndIf
    
    \If{$g = G$}
        \State \textbf{Output} $bestSolution$
        \State \textbf{break}
    \EndIf
    
    \State \textbf{Elitism:} $P_{new}[1] \gets bestSolution$
    
    \For{$i = 2$ to $N$}
        \State Tournament selection to choose parents
        \State Crossover parents to create offspring
        \State Mutation of offspring
        \State $P_{new}[i] \gets$ offspring
    \EndFor
    
    \State Update population: $P \gets P_{new}$
\EndFor
\State \textbf{return} $bestSolution$ with $bestFitness$
\end{algorithmic}
}
\end{algorithm}

\vspace{0em}
\begin{center}
\textbf{Algorithmic Description}
\end{center}
\end{minipage}
\clearpage

\noindent plane and their dimensions for antenna size reduction, while maintaining $S_{11}$ performance at 2.4~GHz, 3.6~GHz, and 5.2~GHz. Unlike traditional Pareto-based methods that maintain multiple solutions, the proposed MOGA converts a multi-objective problem into a single fitness function by summing the $S_{11}$ values at these frequencies, thereby enabling convergence toward a single optimal solution. 

The algorithm incorporates elitist selection to preserve the best individuals across generations, combined with binary tournament selection for parent selection and standard crossover/mutation operations for offspring generation. This approach provides superior convergence characteristics for antenna optimization problems, where a single best-compromise solution is preferred over a Pareto front, particularly when addressing conflicting objectives, such as minimizing return loss across multiple frequencies while simultaneously reducing antenna dimensions and optimizing CSRR placement within the ground plane. 

The weighted-sum strategy facilitates the fine-tuning of the relative importance assigned to different frequency bands and design constraints, rendering it particularly suitable for practical antenna design applications where engineering trade-offs must be explicitly defined. The algorithm flowchart and details are presented below.
\section{Results and Discussion}
The automated design methodology for microstrip patch antennas with CSRR was implemented following the framework proposed by \cite{ref32}. This approach investigates an automated antenna design by optimizing the dimensions of the patch, substrate, ground plane, and CSRR etched into the ground plane. Employing optimization techniques such as MOGA enables the determination of optimal CSRR positioning and dimensions, thereby reducing the antenna size while maintaining the performance metric $S_{11} < -10$ dB. In this study, we converted the multi-objective antenna optimization problem into a single-objective formulation using weighted-sum scalarization, enhanced with constraint-repair mechanisms and elitist preservation strategies.

\subsection{Experimental Setup.}
Figure 4(a) shows that the patch geometry is based on a rectangular structure with a centrally located narrow slot extending toward the feed line. This slot divides the lower portion of the patch into two equal sections, effectively modifying the surface current distribution and enabling a multi-band operation. Figure 5(b) illustrates the ground plane, which incorporates a DGS that behaves similar to a CSRR formed by two concentric circular slots. Each ring contains a narrow gap positioned on the opposite side to maintain symmetry. When excited by an electromagnetic field, currents circulate around the rings of the CSRR structure in response to incident electromagnetic waves. These circulating currents generate a magnetic field that manifests as an inductive effect. The metallic rings, separated by dielectric gaps, form capacitive elements where the electric field concentrates, creating capacitance between the adjacent conductive surfaces. The gaps in the rings are critical features in which the electric fields are strongly concentrated, producing significant capacitive effects.

\begin{figure*}[t!]
\centering
\includegraphics[width=0.7\textwidth]{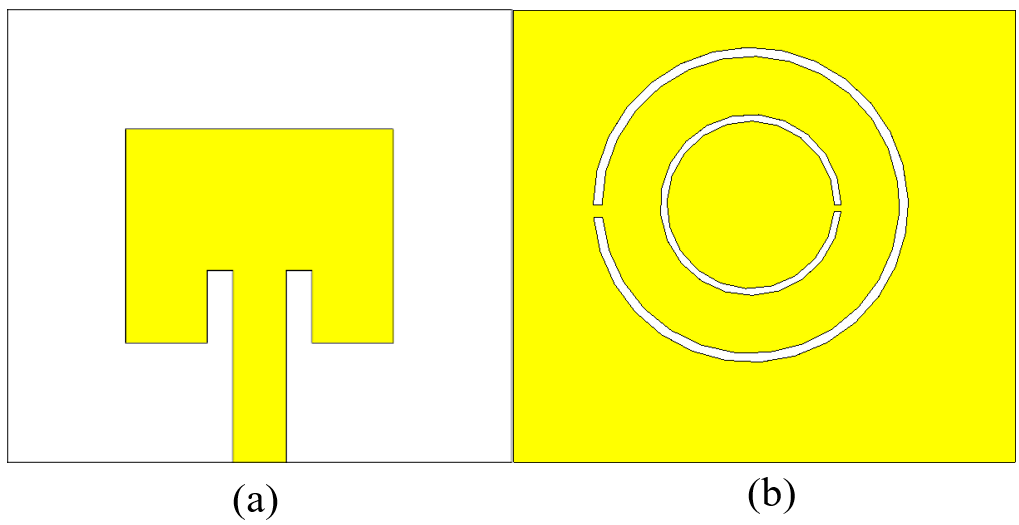}
\caption{Geometry of the proposed antenna: (a) top view and (b) bottom view}
\label{fig1}
\end{figure*}

\begin{figure*}[t!]
\centering
\includegraphics[width=0.7\textwidth]{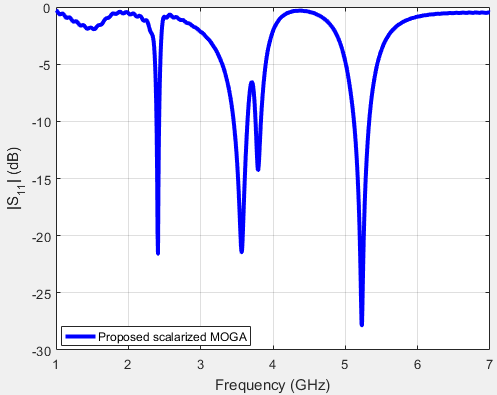}
\caption{Reflection Coefficient $S_{11}$ of the Proposed Antenna}
\label{fig1}
\end{figure*}

Additionally, Figure~5 illustrates the simulated return loss $S_{11}$ of the proposed antenna and the algorithm over the frequency range of 0~GHz to 7~GHz. The blue curve shows the tri-band operation encompassing all the three target frequencies. The deepest resonances occur at approximately 2.4~GHz and 5.2~GHz, with values of $-21.56$~dB and $-27.69$~dB, respectively, making them particularly well-optimized for these bands. At 3.6~GHz, the algorithm achieves a minimum $S_{11}$ of $-16.60$~dB, demonstrating stable resonance and effective power transfer for 5G applications. The design successfully covered multiple wireless communication standards, including Wi-Fi, 5G, and IoT applications.

Figure~6 illustrates the E- and H-plane 2D polar plots of the far-field radiation pattern of the antenna. Figure~6(a) shows the E-plane in the vertical (elevation cut) plane, where $\phi = 0^\circ$ and $\phi = 180^\circ$ are fixed and $\theta$ varies from $0^\circ$ to $360^\circ$. The vertical axis represents the angle of radiation and the horizontal axis represents the maximum gain. The orange curve shows the lowest gain, the green curve shows the intermediate gain, and the blue curve shows the largest pattern with the highest gain. Figure~6(b) shows the H-plane in the horizontal direction (azimuth cut), illustrating how the antenna radiates in a plane perpendicular to the electric field polarization. The H-plane is taken at $\theta = 90^\circ$ using polar coordinates with an azimuth angle $\phi$ ranging from $0^\circ$ to $360^\circ$. All three patterns exhibited multiple lobes distributed around the azimuth, indicating a complex radiation behavior in the H-plane. The pattern at 5.2~GHz shows different directional characteristics compared with those at the other two frequencies.

\begin{figure*}[t!]
\centering
\includegraphics[width=0.7\textwidth]{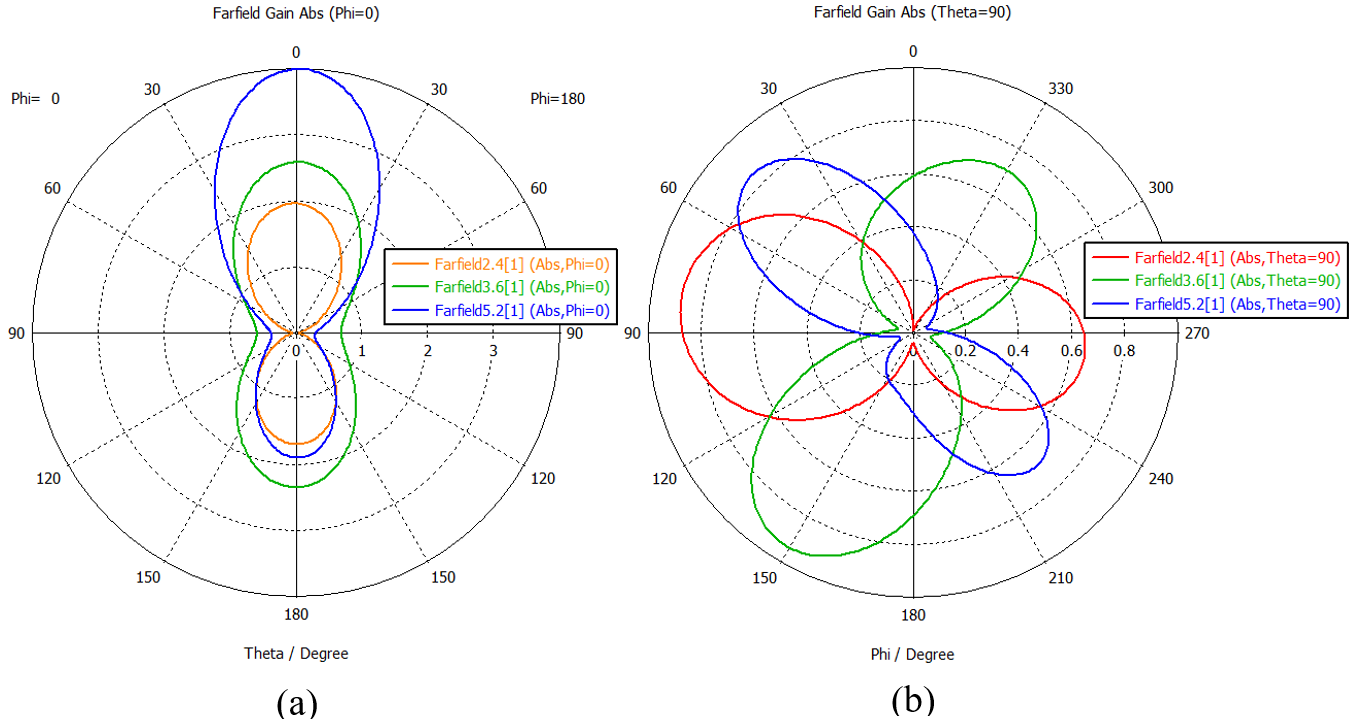}
\caption{Measured farfield radiation patterns: (a) E-plane ($\phi=0°$) and (b) H-plane ($\theta=90°$) at 2.4, 3.6, and 5.2~GHz}
\label{fig1}
\end{figure*}

\subsection{Influence of Various MOGA Performance Metrics}
In this study, we evaluate the performance metrics of different MOGA. We employ generational distance (GD) \cite{ref33} and inverted generational distance (IGD) \cite{ref34} as performance metrics in our experiments. Both metrics can be explained as follows: the variable $d$ represents the Euclidean distance between design points in the multi-dimensional parameter space (patch length, width, slot dimensions, CSRR radius, gap size, etc.). Variable $A$ represents the current antenna design, containing the actual structural parameters (dimensions, material properties, and geometry), and $P^*$ represents the optimal antenna design, containing the ideal parameter values that the optimization algorithm is attempting to reach. The metrics GD and IGD respectively defined as:

\begin{equation}
\text{GD}(A, P^*) = \frac{1}{|A|} \sqrt{\sum_{i=1}^{|A|} d_i^2}
\end{equation}

\begin{equation}
\text{IGD}(A, P^*) = \frac{1}{|P^*|} \sqrt{\sum_{i=1}^{|P^*|} d_i^2}
\end{equation}

where $|A|$ denotes the cardinality of set $A$, $|P^*|$ denotes the cardinality of the optimal Pareto front $P^*$, and $d_i$ denotes the Euclidean distance from a solution point to the nearest point in the reference set.

Figure 7 illustrates the reflection coefficient versus frequency response, demonstrating the comparative performance of the six different optimization approaches across the 1--7~GHz frequency range. The blue curve represents the Pareto-based GA, which exhibits resonances at approximately 2.4~GHz, 3.6~GHz, and 5.2~GHz, with moderate return loss values around $-10$ to $-20$~dB. This algorithm provides multi-band operation, but with a limited impedance matching depth. The non-dominated sorting GA variants (I, II, III; green, black, and gray curves) display varying degrees of optimization success, with different resonance frequencies and bandwidth characteristics. These approaches showed good performance with less consistency across all bands. The pink curve represents the SPEA, which exhibits deep resonance at certain frequencies but shows narrow bandwidths and may not cover all the desired frequency bands. The red curve represents our proposed algorithm, which shows excellent matching characteristics with $S_{11}$ values well below $-10$~dB at the desired frequencies of 2.4~GHz, 3.6~GHz, and 5.2~GHz.

\begin{figure*}[t!]
\centering
\includegraphics[width=0.7\textwidth]{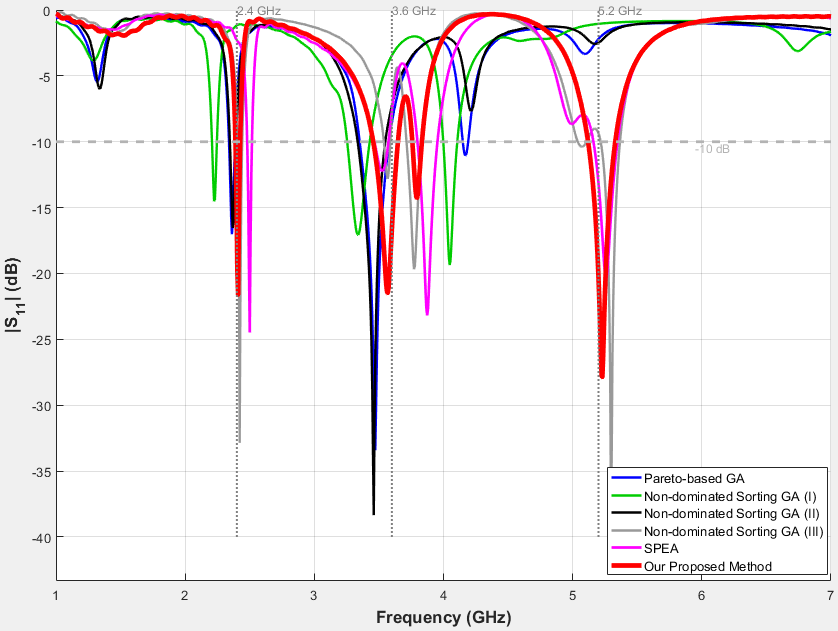}
\caption{Performance comparison of multi-objective genetic algorithm variants for the reflection coefficient $S_{11}$}
\label{fig1}
\end{figure*}

Figure 8 shows the voltage standing wave ratio (VSWR) versus the frequency response. This figure illustrates the impedance-matching performance of six optimization approaches across a frequency range of 1--7~GHz frequency range. Lower VSWR values (ideally, $< 2$) indicate better impedance matching and efficient power transfer. The Pareto-based GA showed stable performance with minimal fluctuations, indicating reasonable impedance matching throughout the spectrum. This algorithm demonstrates acceptable VSWR performance with $1.8 < \text{VSWR} < 2.2$ at target frequencies of 2.4, 3.6, and 5.2~GHz. The non-dominated sorting GA variants (I, II, III; green, black, and gray curves) demonstrate good VSWR performance with values in the range $1.2 < \text{VSWR} < 1.8$ across most desired frequencies. The SPEA displays moderate VSWR performance with a significant peak around 4.5~GHz, where the VSWR reaches approximately 55 at the target frequencies. Our proposed algorithm exhibits distinctive characteristics with an extremely low VSWR ($< 1.5$), indicating very rich impedance matching at the target frequencies.

\begin{figure*}[t]
\centering
\includegraphics[width=0.7\textwidth]{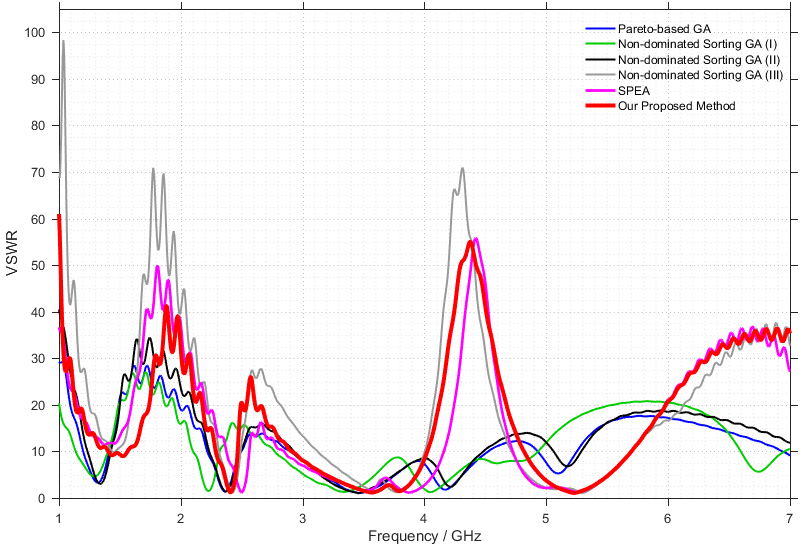}
\caption{Performance comparison of multi-objective genetic algorithm variants for the VSWR}
\label{fig1}
\end{figure*}

Figure 9 illustrates the GD metric, which measures the proximity of the obtained solutions to the true Pareto-optimal front, where lower GD values indicate superior convergence toward optimal solutions. The Pareto GA exhibits an unstable pattern, with peaks at generations 3-9 ($\text{GD} \approx 0.35$), indicating poor consistency and difficulty in maintaining proximity to the optimal front. By Generation 10, the GD converges to approximately 0.27 , which represents a relatively poor performance compared to the other methods. The NSGA-I begins with high GD values ranging from 0.40 to 0.44 throughout all generations, demonstrating the poorest overall convergence performance. In contrast, NSGA-II exhibits good and stable convergence characteristics. Starting at $\text{GD} \approx 0.42$ (generation 1), the algorithm demonstrates consistent monotonic improvement, steadily decreasing to $\text{GD} \approx 0.37$ by generation 10 . The NSGA-III displayed excellent convergence with moderate stability. Beginning at $\text{GD} \approx 0.29$, the algorithm initially shows a slight deterioration to $\text{GD} \approx 0.23$ around generation 5, then stabilizes at approximately $\text{GD} \approx 0.22$ by generation 8, and further improves to $\text{GD} \approx 0.19$ at generation 10. Finally, the SPEA exhibits a distinctive convergence pattern but experiences premature termination at generation 7, preventing complete performance evaluation. Starting at $\text{GD} \approx 0$, the algorithm shows an initial increase to $\text{GD} \approx 0.37$ at generation 2, followed by rapid improvement. the SPEA demonstrates dramatic but unstable convergence, reaching $\text{GD} \approx 0.1$ by generation 5, and approximately $\text{GD} \approx 0.35$ at generation 7 .

\begin{figure*}[b!]
\centering
\includegraphics[width=0.7\textwidth]{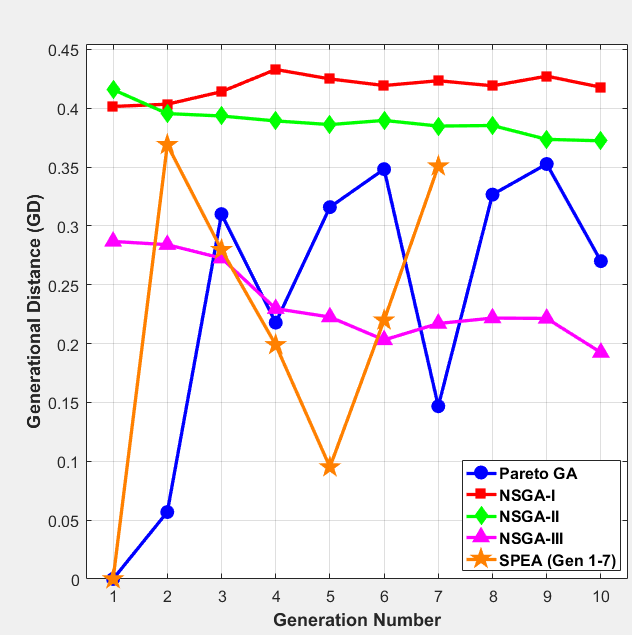}
\caption{Generational distance (GD) Performance of Multi-Objective Genetic Algorithm Variants}
\label{fig1}
\end{figure*}

Figure 10 illustrates the IGD metric, which evaluates convergence quality, and solution diversity. The Pareto GA displays highly erratic and unstable behavior with severe oscillations throughout the optimization process. The algorithm oscillates between $\text{IGD} \approx 0.13$ and $0.58$ , with a final value of approximately $\text{IGD} \approx 0.35$ at generation 10 . This highly unstable pattern suggests that the algorithm struggles to maintain convergence and diversity simultaneously. 

\begin{figure*}[t!]
\centering
\includegraphics[width=0.7\textwidth]{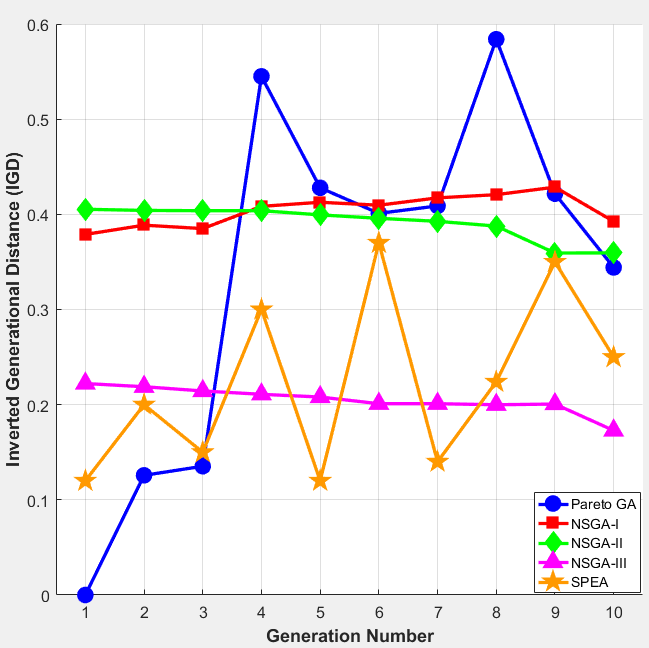}
\caption{Inverted generational distance (IGD) Performance of Multi-Objective Genetic Algorithm Variants}
\label{fig1}
\end{figure*}

The three Non-dominated sorting genetic algorithm variants demonstrated distinct performance characteristics in IGD convergence. NSGA-I exhibited the most predictable behavior, with moderate IGD values ranging from 0.38 to 0.45 , maintaining consistent stability throughout all generations, despite not achieving the lowest values. NSGA-II provides excellent stability with a flat profile ($\text{IGD} \approx 0.37$--$0.41$) and sustained low values, demonstrating reliable convergence and solution diversity suitable for practical applications. NSGA-III achieves superior performance with the lowest IGD values (0.18--0.23), combining exceptional stability with continuous improvement to reach $\text{IGD} \approx 0.18$ at generation 10, making it the most effective algorithm for balancing convergence quality and solution diversity in multi-objective antenna optimization. The SPEA shows moderate performance with notable instability and fluctuating behavior. While the SPEA occasionally achieves competitive IGD values, substantial fluctuations suggest difficulty in maintaining stable convergence and diversity across generations.

Figure 11 illustrates the dual performance metrics of the proposed scalarized MOGA, tracking both the convergence speed (blue curve, left axis) and population diversity (red curve, right axis) across 10 generations. The algorithm demonstrated a rapid initial convergence phase, followed by stabilization. The final convergence speed of approximately 0.2 at generation 10 indicates that the algorithm has reached a stable optimal solution set with minimal fitness improvements in later generations. Population diversity, measured by the standard deviation of the solutions, shows a dynamic \textit{exploration-exploitation} balance. The consistently high diversity values (mostly above 15.3 on average) throughout the optimization process demonstrate that the proposed MOGA successfully avoids premature convergence and maintains a well-distributed solution set.

\begin{figure*}[t!]
\centering
\includegraphics[width=1\textwidth]{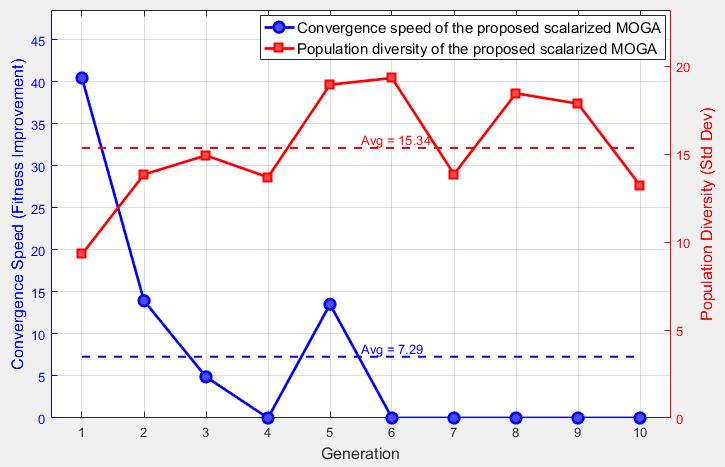}
\caption{Analysis of convergence speed and population diversity of the proposed scalarized multi-objective genetic algorithm for tri-Band antenna optimization}
\label{fig1}
\end{figure*}

Table 3 presents a comparative analysis of the different optimization algorithms. The table displays the optimized antenna parameters, and their corresponding values along with the average size reduction achieved. Additionally, the operating frequencies and associated reflection coefficients $S_{11}$ were reported to evaluate the electromagnetic performance of each optimized design. This comparison demonstrates the effectiveness of the algorithms under consideration for achieving compact antenna structures while maintaining an acceptable impedance matching.

\begin{table}[H]
\makebox[\textwidth][c]{%
\begin{minipage}{1.2\textwidth}
\centering
\caption{Comparison of antenna parameters and performance metrics for different algorithms}
\label{tab:antenna-algorithms-comparison}
\footnotesize
\setlength{\tabcolsep}{4pt}
\renewcommand{\arraystretch}{0.9}
\begin{tabular}{|>{\centering\arraybackslash}m{2cm}|>{\centering\arraybackslash}m{2.5cm}|>{\centering\arraybackslash}m{1.5cm}|>{\centering\arraybackslash}m{3cm}|>{\centering\arraybackslash}m{2cm}|>{\centering\arraybackslash}m{1.2cm}|}
\hline
\textbf{Proposed algorithm} & \textbf{Antenna parameters} & \textbf{Value (mm)} & \textbf{Average antenna size reduction} & \textbf{Frequency (GHz)} & \textbf{S11 (dB)} \\
\hline
\multirow{8}{*}{\centering PGA} & $W_g$ & 40.70 & \multirow{8}{*}{\centering 19.45\%} & \multirow{3}{*}{\centering 2.4} & \multirow{3}{*}{\centering -16.02} \\
\cline{2-3}
 & $L_g$ & 35.94 & & & \\
\cline{2-3}
 & $W_p$ & 21.70 & & & \\
\cline{2-3} \cline{5-6}
 & $L_p$ & 16.44 & & \multirow{3}{*}{\centering 3.6} & \multirow{3}{*}{\centering -34.23} \\
\cline{2-3}
 & $R_1$ & 9.70 & & & \\
\cline{2-3}
 & $R_2$ & 8.65 & & & \\
\cline{2-3} \cline{5-6}
 & $R_3$ & 4.95 & & \multirow{2}{*}{\centering 5.2} & \multirow{2}{*}{\centering -1.15} \\
\cline{2-3}
 & $R_4$ & 4.41 & & & \\
\hline
\multirow{8}{*}{\centering NSGA-I} & $W_g$ & 38.88 & \multirow{8}{*}{\centering 7.39\%} & \multirow{3}{*}{\centering 2.4} & \multirow{3}{*}{\centering -1.924} \\
\cline{2-3}
 & $L_g$ & 36.89 & & & \\
\cline{2-3}
 & $W_p$ & 22.66 & & & \\
\cline{2-3} \cline{5-6}
 & $L_p$ & 16.56 & & \multirow{3}{*}{\centering 3.6} & \multirow{3}{*}{\centering -5.367} \\
\cline{2-3}
 & $R_1$ & 12.38 & & & \\
\cline{2-3}
 & $R_2$ & 11.34 & & & \\
\cline{2-3} \cline{5-6}
 & $R_3$ & 6.94 & & \multirow{2}{*}{\centering 5.2} & \multirow{2}{*}{\centering -0.775} \\
\cline{2-3}
 & $R_4$ & 6.36 & & & \\
\hline
\multirow{8}{*}{\centering NSGA-II} & $W_g$ & 40.36 & \multirow{8}{*}{\centering 9.90\%} & \multirow{3}{*}{\centering 2.4} & \multirow{3}{*}{\centering -16.34} \\
\cline{2-3}
 & $L_g$ & 34.05 & & & \\
\cline{2-3}
 & $W_p$ & 20.24 & & & \\
\cline{2-3} \cline{5-6}
 & $L_p$ & 16.96 & & \multirow{3}{*}{\centering 3.6} & \multirow{3}{*}{\centering -33.81} \\
\cline{2-3}
 & $R_1$ & 12.00 & & & \\
\cline{2-3}
 & $R_2$ & 10.82 & & & \\
\cline{2-3} \cline{5-6}
 & $R_3$ & 7.33 & & \multirow{2}{*}{\centering 5.2} & \multirow{2}{*}{\centering -0.98} \\
\cline{2-3}
 & $R_4$ & 5.97 & & & \\
\hline
\multirow{8}{*}{\centering NSGA-III} & $W_g$ & 41.20 & \multirow{8}{*}{\centering 9.11\%} & \multirow{3}{*}{\centering 2.4} & \multirow{3}{*}{\centering -32.81} \\
\cline{2-3}
 & $L_g$ & 34.81 & & & \\
\cline{2-3}
 & $W_p$ & 20.18 & & & \\
\cline{2-3} \cline{5-6}
 & $L_p$ & 16.56 & & \multirow{3}{*}{\centering 3.6} & \multirow{3}{*}{\centering -10.52} \\
\cline{2-3}
 & $R_1$ & 11.41 & & & \\
\cline{2-3}
 & $R_2$ & 11.00 & & & \\
\cline{2-3} \cline{5-6}
 & $R_3$ & 7.69 & & \multirow{2}{*}{\centering 5.2} & \multirow{2}{*}{\centering -10.72} \\
\cline{2-3}
 & $R_4$ & 6.25 & & & \\
\hline
\multirow{8}{*}{\centering SPEA} & $W_g$ & 38.30 & \multirow{8}{*}{\centering 15.96\%} & \multirow{3}{*}{\centering 2.4} & \multirow{3}{*}{\centering -24.94} \\
\cline{2-3}
 & $L_g$ & 36.17 & & & \\
\cline{2-3}
 & $W_p$ & 20.16 & & & \\
\cline{2-3} \cline{5-6}
 & $L_p$ & 17.40 & & \multirow{3}{*}{\centering 3.6} & \multirow{3}{*}{\centering -12.52} \\
\cline{2-3}
 & $R_1$ & 12.20 & & & \\
\cline{2-3}
 & $R_2$ & 11.35 & & & \\
\cline{2-3} \cline{5-6}
 & $R_3$ & 6.94 & & \multirow{2}{*}{\centering 5.2} & \multirow{2}{*}{\centering -10.41} \\
\cline{2-3}
 & $R_4$ & 6.40 & & & \\
\hline
\multirow{8}{*}{\centering \textbf{\begin{tabular}{c}Our\\Proposed\end{tabular}}} & $W_g$ & 38.28 & \multirow{8}{*}{\centering 16.29\%} & \multirow{3}{*}{\centering 2.4} & \multirow{3}{*}{\centering -21.56} \\
\cline{2-3}
 & $L_g$ & 34.10 & & & \\
\cline{2-3}
 & $W_p$ & 20.36 & & & \\
\cline{2-3} \cline{5-6}
 & $L_p$ & 16.22 & & \multirow{3}{*}{\centering 3.6} & \multirow{3}{*}{\centering -16.60} \\
\cline{2-3}
 & $R_1$ & 12.02 & & & \\
\cline{2-3}
 & $R_2$ & 11.34 & & & \\
\cline{2-3} \cline{5-6}
 & $R_3$ & 6.91 & & \multirow{2}{*}{\centering 5.2} & \multirow{2}{*}{\centering -27.69} \\
\cline{2-3}
 & $R_4$ & 6.40 & & & \\
\hline
\end{tabular}
\end{minipage}
}
\end{table}

\subsection{Comparison with Previous Work}
The proposed algorithm demonstrates superior performance compared with existing multi-objective optimization approaches for tri-band antenna design. While the Pareto-based GA exhibited severe instability with erratic convergence patterns and poor final performance and NSGA-I showed consistent but suboptimal results with high $\text{GD} (\approx 0.42)$ and $\text{IGD} (\approx 0.39)$ values, the proposed method achieved significantly better outcomes. Compared to NSGA-II, which provided good stability but moderate convergence quality, and NSGA-III, which achieved the best performance among the conventional methods with $\text{GD} \approx 0.19$ and $\text{IGD} \approx 0.18$. The proposed scalarized MOGA offers a more balanced optimization strategy. The algorithm's rapid convergence speed (average of 7.29) combined with sustained population diversity (average of 15.3) enabled it to effectively explore the solution space while avoiding premature convergence. 

In terms of antenna performance, the proposed method achieves deeper resonances at target frequencies (2.4, 3.6, and 5.2~GHz) with $S_{11}$ values below $-10$~dB and $\text{VSWR} \approx 1.2$--$1.5$, outperforming all the compared algorithms. Although the SPEA showed promising early convergence in GD analysis, its premature termination and instability in IGD metrics make it unreliable, whereas the proposed method maintains consistent performance throughout all generations, establishing it as the most effective and robust solution for multi-band antenna optimization. Table 4 presents a comparative analysis of the proposed antenna and those of previous studies.

\section{Conclusion}

In this study, we developed an automated methodology for the design and optimization of microstrip patch antennas loaded with complementary split-ring resonators (CSRRs) using co-simulation between MATLAB and CST-API. The reference antenna, operating at 5.2~GHz for IoT applications, was designed in three stages: a zero-stage microstrip patch antenna, a first-stage single-ring configuration etched in the ground plane operating at 3.6~GHz, and a second-stage double-ring configuration operating at 5.2~GHz. The antenna was fabricated on a Rogers RT5880 substrate with a dielectric constant of 2.2, a loss tangent of 0.0009 , and a thickness of 1.57~mm. We applied and investigated five constrained multi-objective genetic algorithms (MOGAs), including PGA, NSGA(I-II-III), and PSEA, to optimize the double-ring geometry with a gap position etched in the ground plane. These algorithms served as defective ground structures (DGSs) to reduce the antenna size while maintaining a return loss below $-10$~dB at the desired tri-band frequencies of 2.4~GHz, 3.6~GHz, and 5.2~GHz.

\begin{table}[H]
\makebox[\textwidth][c]{%
\begin{minipage}{1.3\textwidth}
\centering
\caption{Comparison of the Proposed Antenna Design with Previous Studies}
\label{tab:antenna-comparison}
\footnotesize
\setlength{\tabcolsep}{3pt}
\renewcommand{\arraystretch}{1.1}
\begin{tabular}{|>{\centering\arraybackslash}m{1.8cm}|>{\centering\arraybackslash}m{2.8cm}|>{\centering\arraybackslash}m{1.8cm}|>{\centering\arraybackslash}m{1.5cm}|>{\centering\arraybackslash}m{2cm}|>{\centering\arraybackslash}m{3.1cm}|}
\hline
\textbf{Ref} & \textbf{Method} & \textbf{Frequency (GHz)} & \textbf{$S_{11}$ (dB)} & \textbf{Maximum Gain (dBi)} & \textbf{Dimension (mm$^3$)} \\
\hline
\cite{ref35} & PSO & 0.432 & -22 & - & 128$\times$121$\times$2.1 \\
\hline
\cite{ref36} & NSGA-II & 2.45 & -29 & - & 55$\times$62$\times$1 \\
\hline
\cite{ref37} & Standard GA & 2.5 & -25 & 7.5 & 165$\times$85$\times$0.8 \\
\hline
\cite{ref38} & Standard GA & 2.4 & -12 & - & 65$\times$65$\times$1 \\
\hline
\cite{ref39} & GA & 2.6 & -11 & 5.48 & 53$\times$50$\times$1.6 \\
\hline
\cite{ref40} & GA & 2.45 & -41 & 4.8 & 95$\times$125$\times$1.52 \\
\hline
\cite{ref41} & Current distribution within ground plane & 2.42, 5.22, 5.92 & -29.7, -33.9, -24.7 & 8.18 & 55.5$\times$42.75$\times$1.5 \\
\hline
\cite{ref42} & GA+ANN & 2.4, 3.4 & -16, -20 & 2.1 & 42.2$\times$35$\times$1.5 \\
\hline
\cite{ref13} & GA & 4 & -30 & 3.63 & 30.4$\times$25.5$\times$1.6 \\
\hline
\cite{ref43} & PSO+FA & 3.1--14.2 & -10 & - & 21.5$\times$12$\times$1.6 \\
\hline
\cite{ref44} & NSGA-II & 8.3--11 & -30 & 0.65--8.5 & 41.9$\times$41.9$\times$1.6 \\
\hline
\textbf{Our proposed} & \textbf{Scalarized MOGA} & \textbf{2.4, 3.6, 5.2} & \textbf{-21.56, -16.60, -27.69} & \textbf{3.99} & \textbf{38.74$\times$35.39$\times$1.57} \\
\hline
\end{tabular}
\end{minipage}
}
\end{table}

We propose a novel scalarized multi-objective algorithm that transforms multiple objectives into a single fitness function while preserving the input parameters and constraints within the same optimization framework. This algorithm achieved optimal antenna dimensions and ring positions while maintaining return loss values of $-21.56$~dB, $-16.60$~dB, and $-27.69$~dB, with corresponding gains of 1.96~dBi, 2.6~dBi, and 3.99~dBi at the desired frequencies of 2.4~GHz, 3.6~GHz, and 5.2~GHz, respectively.

The optimization process employed a randomly initialized population of $N=10$, where each individual antenna geometry was generated using a uniform random distribution within predefined parameter constraints to ensure physically realizable designs.

\section{Future Work}

Although our proposed algorithm and workflow have been successfully implemented for co-simulation to reduce antenna size, determine the optimal position of the double-ring with gap etched in the ground plane, and maintain return loss below $-10$~dB while maximizing gain at desired frequencies, several areas require further improvement. In the following sections, we discuss the key issues and limitations that remain to be addressed in future work.

The first issue in the optimization process is that the scalarized MOGA cannot achieve or enhance the frequency precision at 3.6~GHz below $-16.60$~dB. The algorithm considers the optimal value at 3.58~GHz to be $-22.30$~dB rather than 3.6~GHz exactly. The second issue concerns the MATLAB--CST co-simulation process, in which it was not possible to save all ASCII tables of antenna radiation patterns at specific frequencies for every individual in the population to build a comprehensive dataset. Only the optimal solution was successfully stored, including the gain, directivity, and 2D polar plot radiation patterns. This issue arises from the lack of proper synchronization between MATLAB and CST at each iteration of the optimization algorithm, whereas synchronization is correctly achieved only for the final optimal solution.

A significant limitation of this study is the absence of a comparison between the simulation and the experimental analysis. Because the study did not fabricate and test the antenna, we could not validate some material properties that are currently unavailable in the simulation. The second constraint is the limited population size and number of generations, which we plan to increase in future studies to achieve more optimal solutions.

In our future work, we will focus on employing alternative algorithms to reduce the number of iterations and enhance optimization performance. These approaches include various fuzzy logic controllers and hidden Markov models with the ultimate goal of implementing reinforcement learning to control the hyperparameters of the genetic algorithm. We believe that this represents the most effective approach for exploiting the optimization landscape and achieving superior antenna designs.

\section*{Declaration of Generative AI and AI-Assisted Technologies}
During the preparation of this manuscript, the author(s) used Claude to assist in generating and refining a paragraph with the objectives of improving academic English, clarity, sentence structure, and consistency in intonation. After using this tool, the author(s) carefully reviewed, edited, and validated the generated content and take(s) full responsibility for the accuracy, originality, and integrity of the published article.

\end{document}